\documentclass[conference]{IEEEtran}
\IEEEoverridecommandlockouts
\usepackage[section]{placeins}

\usepackage{times}
\usepackage{latexsym}
\usepackage{microtype}
\usepackage{inconsolata}

\usepackage[utf8]{inputenc}
\usepackage[T1]{fontenc}
\usepackage[english]{babel}
\usepackage{mathtools}
\usepackage{mleftright}
\usepackage[mathscr]{euscript}
\usepackage{dutchcal}
\usepackage{graphicx,url}
\usepackage{dirtytalk}
\usepackage{dsfont}
\usepackage{setspace}
\usepackage{amssymb}
\usepackage[margin = 1cm]{caption}
\usepackage{sidecap}
\usepackage{float}
\usepackage{amsmath}
\usepackage{amsthm}
\usepackage{color,epstopdf}
\usepackage[noend]{algpseudocode}
\usepackage{hyperref}
\usepackage{algorithm,algpseudocode}
\usepackage{breqn}

\usepackage{epsfig}
\usepackage{tabu}
\usepackage{arydshln}
\usepackage{xcolor}
\usepackage{mathrsfs}
\usepackage{caption}
\usepackage{subcaption}

\usepackage{enumitem}

\DeclarePairedDelimiterX{\expectarg}[1]{[}{]}{%
  \ifnum\currentgrouptype=16 \else\begingroup\fi
  \activatebar#1
  \ifnum\currentgrouptype=16 \else\endgroup\fi
}

\newcommand{\LinesNumbered}{
  \setboolean{algocf@linesnumbered}{true}%
  \renewcommand{\algocf@linesnumbered}{\everypar={\nl}}}%
\let\oldnl\nl
\newcommand{\nonl}{\renewcommand{\nl}{\let\nl\oldnl}}

\newcommand{\innermid}{\nonscript\;\delimsize\vert\nonscript\;}
\newcommand\tab[1][1cm]{\hspace*{#1}}

\newcommand{\activatebar}{%
  \begingroup\lccode`\~=`\|
  \lowercase{\endgroup\let~}\innermid 
  \mathcode`|=\string"8000
}

\algnewcommand{\Inputs}[1]{%
  \State \textbf{Inputs:}
  \Statex \hspace*{\algorithmicindent}\parbox[t]{.8\linewidth}{\raggedright #1}
}
\algnewcommand{\Initialize}[1]{%
  \State \textbf{Initialize:}
  \Statex \hspace*{\algorithmicindent}\parbox[t]{.8\linewidth}{\raggedright #1}
}
\algnewcommand{\Data}[1]{%
  \State \textbf{Data:}
  \Statex \hspace*{\algorithmicindent}\parbox[t]{.8\linewidth}{\raggedright #1}
}

\graphicspath{{images/}} 

\title{Topic Analysis with Side Information: A Neural-Augmented LDA Approach}
\author{
  \textbf{Biyi Fang\textsuperscript{1}}, 
  \textbf{Truong Vo\textsuperscript{1}},
  \textbf{Kripa Rajshekhar\textsuperscript{2}},
  \textbf{Diego Klabjan\textsuperscript{1}}\\
  \textsuperscript{1}Northwestern University, \textsuperscript{2}Metonymize \\
  \texttt{biyifang2021@u.northwestern.edu, truongvo2025@u.northwestern.edu} \\
  \texttt{kripa@metonymize.com, d-klabjan@northwestern.edu} 
}
\begin{document}
\maketitle
\begin{abstract}
Traditional topic models such as Latent Dirichlet Allocation (LDA) have been widely used to uncover latent structures in text corpora, but they often struggle to integrate auxiliary information such as metadata, user attributes, or document labels. These limitations restrict their expressiveness, personalization, and interpretability. To address this, we propose nnLDA—a neural-augmented probabilistic topic model that dynamically incorporates side information through a neural prior mechanism. nnLDA models each document as a mixture of latent topics, where the prior over topic proportions is generated by a neural network conditioned on auxiliary features. This design allows the model to capture complex, nonlinear interactions between side information and topic distributions that static Dirichlet priors cannot represent. We develop a stochastic variational Expectation–Maximization algorithm to jointly optimize the neural and probabilistic components. Across multiple benchmark datasets, nnLDA consistently outperforms LDA and Dirichlet-Multinomial Regression in topic coherence, perplexity, and downstream classification. These results highlight the benefits of combining neural representation learning with probabilistic topic modeling in settings where side information is available.
\end{abstract}


\section{Introduction}

The exponential growth of user-generated content—ranging from news articles and reviews to blogs and discussion forums—has intensified the demand for interpretable models capable of uncovering meaningful themes in unstructured text. Latent Dirichlet Allocation (LDA)~\cite{blei2003latent} has long served as a cornerstone in this field, providing a principled probabilistic framework that balances statistical rigor with semantic interpretability. However, classical topic models such as LDA rely on fixed priors and operate exclusively on textual data, limiting their ability to incorporate valuable auxiliary information such as document metadata, user attributes, or categorical labels.

Existing efforts to enhance LDA with side information typically fall into two categories. Downstream models treat auxiliary data as targets to be predicted jointly with topics, while upstream models use such data as conditioning variables that shape the topic-generation process. Although these approaches have improved performance in certain settings, many still depend on linear assumptions or static priors, as exemplified by Dirichlet-Multinomial Regression (DMR)~\cite{Mimno2008TopicMC}. These limitations constrain their flexibility and ability to capture complex relationships between side information and latent topics.

To address these shortcomings, we propose nnLDA (Neural Network–augmented Latent Dirichlet Allocation)—a neural-enhanced probabilistic topic model that dynamically integrates side information through learned priors. nnLDA bridges the gap between probabilistic topic modeling and neural representation learning by employing a feedforward neural network to generate document-specific Dirichlet priors conditioned on auxiliary features. This mechanism enables nnLDA to adapt topic distributions at the document level, capturing subtle and nonlinear dependencies that traditional LDA cannot express. The model preserves the generative structure of LDA while introducing a neural prior module that learns context-aware topic mixtures.

We develop a stochastic variational Expectation–Maximization (EM) algorithm for efficient optimization, where the E-step infers topic assignments for each word and the M-step jointly updates both the topic-word distributions and neural network parameters. Furthermore, we provide a theoretical analysis demonstrating that nnLDA achieves a variational bound at least as tight as that of standard LDA.

Extensive experiments on multiple benchmark datasets show that nnLDA consistently outperforms both LDA and DMR in topic coherence, perplexity, classification accuracy, and generative quality. In summary, our contributions are threefold:
\begin{itemize}
\item We introduce a neural-augmented topic model that conditions document-specific priors on side information;
\item We develop a variational EM framework for joint optimization of neural and probabilistic components;
\item We empirically demonstrate that nnLDA enhances both interpretability and predictive performance across diverse datasets.
\end{itemize}

\section{Related Work}
There are a large amount of extensions of the plain LDA model, however, a full retrospection of this immense literature exceeds the scope of this work. In this section, we state several kinds of variations of LDA which are most related to our new model and interpret the relationships among them.

\textit{\textbf{LDA:}}
The original LDA model has been extensively applied in text analysis~\cite{blei2003latent,Wang2020KeywordbasedTM}, image modeling~\cite{Li2005ABH}, and network analysis~\cite{Airoldi2008MixedMS}, due to its simplicity, low-dimensional representations, and interpretable topic structures. However, adapting LDA to new tasks typically requires re-deriving its inference procedure. To improve scalability and flexibility, Srivastava and Sutton~\cite{srivastava2017autoencoding} introduced Neural Variational LDA (NVLDA), which replaces the Dirichlet prior with a Logistic-Normal distribution to enable amortized inference. RollingLDA~\cite{riegeretal2021} supports sequential updates for streaming data, while Optimized LDA (OLDA)~\cite{olda2024} employs hyperparameter tuning to enhance topic quality. Despite these advances, conventional LDA and its direct variants operate solely on textual inputs, neglecting valuable auxiliary information such as user or document metadata. In contrast, our proposed nnLDA integrates such side data through a neural network that adaptively learns document-specific priors, allowing the model to capture richer and more nuanced topic–feature relationships.

\textit{\textbf{Downstream Topic Models:}} 
Downstream extensions of LDA jointly generate both the text and its associated side information from shared latent topics. In these models, each topic is associated with two distributions—one over words and another over side attributes—and learning proceeds by maximizing their joint likelihood. Representative examples include Correspondence LDA (Corr-LDA)~\cite{Blei2003ModelingAD}, the Mixed-Membership Model for Authorship~\cite{Erosheva2004MixedmembershipMO}, the Group-Topic Model~\cite{Wang2005GroupAT}, Topics over Time (TOT)~\cite{Wang2006TopicsOT}, Maximum Entropy Discrimination LDA (MedLDA)~\cite{Zhu2012MedLDAMM}, and the Term–URL Model (TUM)~\cite{jiang2013beyond}.
For instance, TUM models search engine queries by separately generating terms and URLs, which increases computational complexity due to distinct generative paths. Another influential variant is Supervised LDA (sLDA)~\cite{Blei2007SupervisedTM}, which incorporates response variables such as ratings by modeling them through a generalized linear model (GLM). However, such approaches require explicit specification of link and dispersion functions, limiting their flexibility to only a small number of side-data types. Our nnLDA framework, in contrast, sidesteps these constraints by using a neural prior that learns nonlinear mappings from arbitrary auxiliary features to document-specific topic priors.

\textit{\textbf{Upstream Topic Model:}}
Upstream models differ fundamentally from downstream ones by conditioning topic generation on observed side information, thus maximizing conditional rather than joint likelihood. Classic examples include Discriminative LDA (DiscLDA)~\cite{LacosteJulien2008DiscLDADL}, Scene Understanding Models~\cite{Sudderth2005LearningHM}, and the Author–Topic Model~\cite{RosenZvi2004TheAM}, where each word is generated by first selecting an author and then sampling a topic from that author’s distribution. Subsequent extensions~\cite{RosenZvi2004TheAM,McCallum2007TopicAR,Dietz2007UnsupervisedPO} allow topic mixtures per document but are typically limited to a single modality (e.g., labels or ratings) and cannot simultaneously accommodate continuous and categorical side data.

Later approaches such as Dirichlet–Multinomial Regression (DMR)~\cite{Mimno2008TopicMC} and Collective Supervision Models~\cite{benton2016collective} address this limitation partially by projecting side information into topic priors using a dot-product parameterization. However, such linear mappings remain restrictive. Our nnLDA generalizes these ideas by introducing a neural network that learns nonlinear, data-driven transformations from side information to prior parameters. This enables the model to flexibly combine heterogeneous modalities and better adapt topic distributions to contextual cues, improving both inference and interpretability.

\section{Algorithms}
We first establish notation and define the setting. Throughout this section, we use the terminology of \textit{text collections}—such as ``words,'' ``documents,'' and ``corpora''—to make the concepts more intuitive. Similar to standard LDA, however, the proposed nnLDA model is not limited to textual datasets and can also be applied to other modalities such as images or networks.

\begin{itemize}
    \item \textbf{Word:}  
    A \textit{word} is an element from a vocabulary of size $V$, indexed by $\{1, \ldots, V\}$, and represented using one-hot encoding. Specifically, the $v$-th word is represented by a $V$-dimensional vector $w$, where $w^v = 1$ and $w^u = 0$ for all $u \neq v$.

    \item \textbf{Document:}  
    A \textit{document} is a collection of $N$ words, denoted as
    \[
    d = \mathbf{w} = \{w_1, w_2, \ldots, w_N\}.
    \]
    When side information is available, the document is expressed as
    \[
    d = (\mathbf{w}, \mathbf{s}) = (\{w_1, w_2, \ldots, w_N\}, (s_1, s_2, \ldots, s_q)),
    \]
    where $\mathbf{s} \in \mathbb{R}^q$ contains $q$ auxiliary features, which may be categorical or continuous.

    \item \textbf{Corpus:}  
    A \textit{corpus} is a collection of $M$ documents. For text-only data,
    \[
    D = \{\mathbf{w}_1, \mathbf{w}_2, \ldots, \mathbf{w}_M\},
    \]
    and when side information is included,
    \[
    (D, S) = \{(\mathbf{w}_1, \mathbf{s}_1), (\mathbf{w}_2, \mathbf{s}_2), \ldots, (\mathbf{w}_M, \mathbf{s}_M)\}.
    \]
\end{itemize}

The primary objective of nnLDA is to learn a probabilistic model of the corpus that, by leveraging high-level summarization from side data, assigns high probability not only to the observed documents but also to similar, unobserved ones sharing comparable side-information characteristics.

\subsection{Generative Model}

We now describe the nnLDA generative process for a document $d = (\mathbf{w}, \mathbf{s})$ containing $N$ words and side information $\mathbf{s}$. The process proceeds as follows:

\begin{algorithm}[H]
\caption{Generative Process of nnLDA}
\begin{algorithmic}[1]
\State Choose $N \sim \text{Poisson}(\xi)$
\State Choose $\mathbf{s} \sim \mathcal{N}(\mu, \sigma^2 I)$
\State Compute $\alpha_d \gets g(\gamma; \mathbf{s})$
\State Draw $\theta \sim \text{Dir}(\alpha_d)$
\For{each of the $N$ words $w_n$}
    \Statex \quad (a) Draw topic $z_n \sim \text{Multinomial}(\theta)$
    \Statex \quad (b) Draw word $w_n$ from $p(w_n \mid z_n, \beta)$
\EndFor
\end{algorithmic}
\end{algorithm}

Here, ``Poisson'' and ``Dir'' denote the Poisson and Dirichlet distributions, respectively. In Step 3, $g$ denotes a parametric neural network that maps side data $\mathbf{s}$ to document-specific Dirichlet parameters $\alpha_d$.  
The model therefore contains two sets of trainable parameters:
\begin{itemize}
    \item $\gamma$: the parameters of the neural network $g(\cdot)$ for encoding side information;
    \item $\beta$: the topic–word distribution.
\end{itemize}
Additionally, there are three hyperparameters:
\begin{itemize}
    \item $\mu$ and $\sigma^2$: the mean and variance of the Gaussian prior for side data $\mathbf{s}$;
    \item $K$: the number of topics (not explicitly shown in the process).
\end{itemize}

Step 1 determines the document length independently of the remaining steps.  
Step 2 samples the side-data representation $\mathbf{s}$ from a normal distribution.  
In Step 3, the neural network $g(\gamma; \mathbf{s})$ produces a document-specific Dirichlet prior $\alpha_d$, which governs the topic proportion $\theta$ drawn in Step 4.  
Finally, Steps 5(a)–(b) follow the standard LDA process: a topic $z_n$ is drawn for each word, and then the word $w_n$ is generated from the corresponding topic–word distribution $\beta$.

\subsection{Analysis}

A Dirichlet random vector $\theta = (\theta_1, \theta_2, \ldots, \theta_K)$ has the following probability density:
\[
p(\theta \mid \alpha) = 
\frac{\Gamma\left(\sum_{i=1}^{K}\alpha_i\right)}
{\prod_{i=1}^{K}\Gamma(\alpha_i)}
\theta_1^{\alpha_1 - 1} \cdots \theta_K^{\alpha_K - 1},
\]
where $K$ is the number of topics, $\alpha$ is the prior, and $\theta$ lies on the $(K-1)$-simplex.  

Accordingly, the conditional distribution of a document $d = (\mathbf{w}, \mathbf{s})$ under nnLDA can be expressed as:
\begin{align*}
P_1(\mathbf{w} \mid \mu, \sigma, \gamma, \beta)
&= \int p(\theta \mid \mu, \sigma, \gamma)
\left(
\prod_{n=1}^{N} \sum_{i=1}^{K}
\prod_{j=1}^{V} (\theta_i \beta_{ij})^{w_n^j}
\right) \mathrm{d}\theta,
\end{align*}
where $p(\theta \mid \mu, \sigma, \gamma) = p(\theta \mid g(\gamma; \mathbf{s})) = p(\theta \mid \alpha_d)$.

The nnLDA model is a hierarchical probabilistic graphical model with three levels:
\begin{itemize}
    \item \textbf{Corpus-level parameters:} $\mu$, $\sigma$, $\gamma$, and $\beta$, sampled once for the entire corpus;
    \item \textbf{Document-level variables:} $\alpha_d$ and $\theta_d$, sampled once per document;
    \item \textbf{Word-level variables:} $w_{dn}$ and $z_{dn}$, sampled once per word.
\end{itemize}

Compared to standard LDA, nnLDA introduces an additional neural component $g(\gamma; \cdot)$ that generates document-level priors $\alpha_d$. Because nnLDA generalizes LDA, it is expected to achieve at least comparable, and often superior, likelihood. Under the assumption that $g$ possesses \textit{finite sample expressivity}~\cite{Yun2019SmallRN}, nnLDA can approximate or replicate the optimal Dirichlet priors $\alpha^*$ learned by LDA, ensuring that
\[
P_1(D \mid \mu^*, \sigma^*, \gamma^*, \beta^*) 
\geq 
P_2(D \mid \alpha^*, \beta^*),
\]
where $P_1$ and $P_2$ denote the optimized likelihoods under nnLDA and LDA, respectively.

\section{Experimental Study}
In this section, we compare the nnLDA model with standard LDA and the DMR model introduced in \cite{blei2003latent} and \cite{Mimno2008TopicMC}, respectively. We conduct experiments on five different-size datasets among which one is a synthetic dataset and the remaining four are real-world datasets. For these datasets, we study the performance of topic grouping, perplexity, classification and comment generation for nnLDA, plain LDA and DMR models. For each of the tasks, some datasets are not eligible to be examined due to lack of information. The synthetic dataset is publically available at \url{https://github.com/biyifang/nnLDA/blob/main/syn_file.csv} while the real-world datasets are proprietary.
\subsection{Datasets and Training Details}
The first dataset we use is a synthetic dataset of 2,000 samples. Each sample contains a customer’s feedback with respect to his or her purchase along with the characteristics of the product. More precisely, there are two different categories, which are product and description. In the product category, it can either be TV or burger; similarly, in the description category, the word can either be price or quality. In order to generate comments, we assign a bag of words to each combination of product and description as shown in Table 
After randomly selecting one category combination from the four combinations, a comment is generated containing at least one word and at most five words with an average 2.97 words, by selecting a certain number of words at random from the corresponding bag. 

The second dataset is a real-world dataset, PTS for short, which has 795 samples. Each sample contains a customer’s short feedback and rating with respect to his or her purchase along with the characteristics of the product. Additionally, the category (side data) selected for nnLDA corresponds to sectors, which are generalizations of products. In this dataset, there is only only 1 word in the shortest comment, while the longest comment in the dataset contains 49 words. Overall, the average length of the comments is 10.6 words. For example, a customer, who bought a product belonging to sector Baby, leaves a comment \say{Cheap\& Soft} with a rating of 3.

The third dataset WIP is a medium-size dataset with 3,451 samples. Each sample contains a customer’s short feedback and rating with respect to his or her purchase along with the characteristics of the product. The sector attribution is again side data when training models with one feature. The other attribution counted for models with two features is channel. The most concrete comment in the dataset has 138 words, while the briefest comment has only 1 word. In the meanwhile, the average length of the comments in the dataset is 8.9 words.

DCL is another medium-size dataset of 5,427 samples. Different from the PTS and WIP datasets, each sample in DCL contains a customer’s long feedback and rating with respect to his or her purchase along with the characteristics of the product. Additionally, the side data selected for nnLDA corresponds to groups of products. The smallest number of words for a comment in this dataset is 1, while the largest is 988. Overall, the average length of the comments is 61.7 words. A short sample comment is \say{quick points that will be all that matters to a buyer wanting accurate metrics to buy by tinny sound but plenty of audio hookups.}

The last dataset is RR, which has 100,000 samples, from which we randomly select 10,000 samples. Each sample contains a customer’s feedback with respect to his or her purchase along with the characteristics of the product. Additionally, the side data for nnLDA corresponds to the category, which can be grocery, health and personal care, furniture, kitchen, etc. The longest comment has 418 words while the shortest comment has only 1 word as the previous datasets, and the average length of the comments is 69.5 words. A short sample comment reads \say{great tasting oil and made the most excellent gluten free chocolate cake.}
\ref{tb1:syn}. 
\begin{table}[t]
\resizebox{\columnwidth}{!}{
\begin{tabular}{|l|l|}
\hline
Category combination & \multicolumn{1}{c|}{Bag of words}                                                                                                  \\ \hline
(burger, price)      
& \begin{tabular}[c]{@{}l@{}}value, pricey, ouch, steep,\\
cheap, value, reason, accept,\\
unreason, unacceptable\end{tabular}         \\ \hline
(burger, quality)    
& \begin{tabular}[c]{@{}l@{}}nasty, fantastic, delicious, tasty,\\ juicy, unreason, unacceptable,\\
reason, accept, fresh\end{tabular} \\ \hline
(TV, price)          
& \begin{tabular}[c]{@{}l@{}}promotion, affordable, value,\\
increase, expensive, tasty, \\economical, fancy, okay\end{tabular}         \\ \hline
(TV, quality)        & \begin{tabular}[c]{@{}l@{}}fabulous, fantastic, promising,\\ sharp, large, clear, eco friendly, \\fresh, pixilated\end{tabular}      \\ \hline
\end{tabular}
}
\caption{Synthetic Dataset}\label{tb1:syn}
\end{table}

Due to the lack of some information from the certain datasets, we are unable to study all tasks of interest for all of these datasets. For the topic grouping task, we examine the ability of nnLDA, plain LDA and DMR to assign the comments from the same topic group into the same correct topic group. For this task, we only conduct experiments on the synthetic dataset since only the topic groups of the synthetic dataset are clear. For the perplexity task, we compute the logarithm of the perplexity of all the words in the corresponding dataset. We do not study the performance of perplexity for the synthetic dataset since we know the true number of topic groups. For the classification task, we use the probability vector generated by the topic models to predict the rating for that comment. Since the RR dataset does not have ratings, we are unable to examine the classification ability of the topic models on the RR dataset. The last task tests the performance of the topic models on generating new comments. For this task, we only conduct experiments on the two smallest real world datasets since it is of interest how topic models perform given a small number of samples. Table \ref{tb6:tasks} presents the tasks of interest for each dataset.

\begin{table}[t]
\centering
\resizebox{\columnwidth}{!}{
\begin{tabular}{|l|l|l|l|l|}
\hline
Dataset                         & \multicolumn{1}{c|}{Topic grouping} & Perplexity & Classification & \begin{tabular}[c]{@{}l@{}}Comment\\ generation\end{tabular} \\ \hline
Synthetic dataset               & Yes                                 & No         & No           & No  \\ \hline
PTS & No               & Yes        & Yes            & Yes \\ \hline
WIP           & No               & Yes        & Yes            & Yes  \\ \hline
DCL        & No& Yes        & Yes            & No  \\ \hline
RR & No& Yes  & No  & No  \\ \hline
\end{tabular}
}
\caption{Tasks of Interest}\label{tb6:tasks}
\end{table}

For all of these datasets, we employ a two-layer fully connected neural network as $g(\gamma;\cdot)$ in nnLDA. Furthermore, we set the number of neurons to be 20 in the first layer, the number of neurons of the second layer to be the number of topic groups assigned in the beginning and the batch size to be 64. All features of the side data are categorical and are one-hot encoded. Additionally, all weights in $g(\gamma;\cdot)$ are initialized by Kaiming Initialization \cite{he2015delving}. We apply the ADAM algorithm with the learning rate of $0.001$ and weight decay being $0.1$. Meanwhile, we train all the models using EM with exactly the same stopping criteria of stopping E-step and M-step when the average change over the whole training dataset in the expected log likelihood becomes less than 0.01\%. We vary the number of topic groups from 4 to 30. For DMR, we use the same values for the parameters as those in \cite{Mimno2008TopicMC}. All the algorithms are implemented in Python with Pytorch and trained on a single GPU card.

\subsection{Experimental Results}
In this section, we present all the results based on the tasks of interest.

Overall, nnLDA outperforms plain LDA and DMR in all datasets in terms of topic grouping, classification, perplexity and comment generation. Meanwhile, based on the fact that the last two datasets have many more words and more intrinsic concepts in their comments when compared to the first three datasets, nnLDA exceeds the performance of plain LDA and DMR dramatically when a document contains several topics or it is more comprehensive.
\subsubsection{Topic Grouping}
Table \ref{tb7:topic} shows the most frequent 5 words in each topic group generated by plain LDA, DMR and nnLDA when setting the number of topic groups to be 4 in the synthetic dataset. The topic groups generated by plain LDA and DMR are very vague and it is very hard to distinguish which topic group is describing what combination of product and description, while the topic groups given by nnLDA are very distinguishable, i.e. topic group 1 is about (burger, quality), topic group 2 is about (TV, price), topic group 3 is about (TV, quality) and topic group 4 is about (burger, price). It identifies correctly the seed topics. Therefore, nnLDA outperforms plain LDA in grouping.
\begin{table*}[!t]
\centering
\begin{tabular}{|l|l|l|l|}
\hline & \multicolumn{1}{c|}{plain LDA}    &DMR        & nnLDA   \\ \hline
Topic group 1 & \begin{tabular}[c]{@{}l@{}}promising, rebate, sharp,\\ increase, outstanding\end{tabular} & \begin{tabular}[c]{@{}l@{}}pricey, unacceptable,\\ juicy, pixilated  \end{tabular} & \begin{tabular}[c]{@{}l@{}}unreason, unacceptable, \\juicy delicious, nasty\end{tabular}   \\ \hline
Topic group 2 & \begin{tabular}[c]{@{}l@{}}unreason, value, okay,\\ steep, ecofriendly\end{tabular}   &\begin{tabular}[c]{@{}l@{}}ouch, steep, tasty,\\ unreason, promotion\end{tabular}       & \begin{tabular}[c]{@{}l@{}}promotion, increase, tasty,\\ economical, okay\end{tabular}     \\ \hline
Topic group 3 & \begin{tabular}[c]{@{}l@{}}reason, accept, promotion,\\ large, unacceptable\end{tabular}  & \begin{tabular}[c]{@{}l@{}}accept, fantastic, value \\reason, affordable\end{tabular}   & \begin{tabular}[c]{@{}l@{}}fresh, promising, fantastic,\\ large, eco friendly\end{tabular} \\ \hline
Topic group 4 & \begin{tabular}[c]{@{}l@{}}fresh, reason, outstanding,\\  ecofriendly, fantastic\end{tabular}&\begin{tabular}[c]{@{}l@{}}sharp, delicious,\\ accept, fresh, clear\end{tabular}  & \begin{tabular}[c]{@{}l@{}}reason, accept, value,\\ steep, cheap \end{tabular} \\\hline
\end{tabular}
\caption{Top words of groups generated by LDA, DMR and nnLDA}\label{tb7:topic}
\end{table*}

Additionally, based on the top words of topics generated by LDA, DMR and nnLDA, we are able to assign the most related category combination to a comment with respect to a model. Since we have the category combination of each comment, Table \ref{tb10:micro-macro} shows the macro-recall, macro-precision and macro-F1 scores and micro-F1 of LDA, DMR and nnLDA, respectively, when training on the synthetic dataset, and the overall relative improvement of nnLDA. 

\begin{table}[H]
\centering
\resizebox{\columnwidth}{!}{
\begin{tabular}{|l|r|r|r|r|}
\hline
& \begin{tabular}[c]{@{}c@{}}macro\\ precision\end{tabular}
& \begin{tabular}[c]{@{}c@{}}macro\\ recall\end{tabular}
& \begin{tabular}[c]{@{}c@{}}macro\\ F1\end{tabular}
& \begin{tabular}[c]{@{}c@{}}micro\\ F1\end{tabular} 
\\ \hline
LDA        & 0.7238 &0.7272&0.7211& 0.7240\\ \hline
DMR       &0.7238&  0.7460 &     0.7313&  0.7392 \\ \hline
nnLDA     &  0.7401   &  0.7919  &   0.7536 &  0.7905  \\ \hline
\begin{tabular}[c]{@{}c@{}}improvement \\ from LDA\end{tabular} &2.25\%&8.90\% & 4.51\%& 9.19\%\\ \hline
\begin{tabular}[c]{@{}c@{}}improvement \\ from DMR\end{tabular}  
&2.25\%& 6.15\%& 3.05\%&6.94\%\\ \hline
\end{tabular}
}
\caption{Precision, recall and relative improvement of the synthetic dataset generated by LDA, DMR and nnLDA}\label{tb10:micro-macro}
\end{table}

As the table shows, nnLDA outperforms plain LDA and DMR, which implies that nnLDA assigns more samples correctly to the right topic group. Therefore, in general, nnLDA improves the recall, precision and F1 scores. 

In conclusion, nnLDA outperforms standard LDA and DMR in terms of the ability of topic grouping.
\subsubsection{Perplexity}
Figures \ref{fig:proj_perp} and \ref{fig:whatif_perp} represent the log(perplexity) of plain LDA, DMR and nnLDA on the PTS and WIP datasets, respectively. Additionally, in Figure \ref{fig:whatif_perp}, for DMR and nnLDA, we not only conduct experiments on the dataset with the single feature (sector) as the side data, denoted as \say{DMR with single feature} and \say{nnLDA with single feature,} but also on the dataset with two features (sector and channel) as side data, denoted as \say{DMR with two features} and \say{nnLDA with two features,} respectively. The smallest log(perplexity) values generated by plain LDA and DMR are competitive to those of nnLDA for these two datasets. In Figure \ref{fig:proj_perp}, the log(perplexity) value generated by plain LDA increases as the number of topic groups grows, while the log(perplexity) values generated by DMR and nnLDA decrease first and then increase as the number of topic groups increases on the PTS dataset. As it is shown in Figure \ref{fig:whatif_perp}, the log(perplexity) values generated by plain LDA and DMR increase as the number of topic groups grows on the WIP dataset. However, the log(perplexity) values generated by nnLDA decrease first and then increase as the number of topic groups increases on both of the aforementioned datasets. Moreover, we examine DMR and nnLDA models with two features on the WIP dataset, which take both sector and channel attributions as side data into account, in Figure \ref{fig:whatif_perp}. As we can observe, the minimum log(perplexity) generated by nnLDA with two features (sector and channel attributions) is better than that of nnLDA with the single feature (sector attribution), although the optimal number of topic groups occurs at a different point since more side data is provided. Consequently, plain LDA does not learn the datasets, and DMR is able to learn the small datasets. In contrast, nnLDA starts learning the datasets as the log(perplexity) value decreases in the beginning and finds an optimal number of topic groups, then it gets confused since the number of topic groups are more than needed. Furthermore, nnLDA with two features provides better log(perplexity) than nnLDA with the single feature. Therefore, nnLDA is more capable of understanding the datasets; both small and medium-size datasets with short comments. 

\begin{table}[H]
\centering
\resizebox{0.9\columnwidth}{!}{
\begin{tabular}{|l|r|r|r|}
\hline
 running time(s) & plain LDA & DMR    & nnLDA  \\ \hline
PTS          & 3      & 4   & 4   \\ \hline
WIF          & 19     & 24  & 26  \\ \hline
DCL          & 138    & 179 & 191 \\ \hline
\end{tabular}
}
\caption{Running time of different models on different datasets}\label{tb:run_time}
\end{table}

When learning more complex datasets, the advantage of the nnLDA model becomes more pronounced. Figure \ref{fig:TV_2} represents the log(perplexity) of plain LDA, DMR and nnLDA on the DCL dataset. In these figures we observe that the log(perplexity) generated by plain LDA and DMR blows up as the number of topic groups increases, while the log(perplexity) generated by nnLDA decreases first and then increases as the number of topic groups grows. Furthermore, the log(perplexity) values of nnLDA are much smaller than those of plain LDA and DMR. Consequently, nnLDA performs as well as plain LDA and DMR in small and medium-size datasets with short comments, and at the same time, nnLDA explains the datasets better than plain LDA and DMR in medium and large size datasets with long comments. There is also a trade-off between the accuracy and running time as shown in Table \ref{tb:run_time}. In the table, we compare the running time of plain LDA, DMR with one feature and nnLDA with one feature on three different datasets. We observe that nnLDA spends more time than both DMR and plain LDA on training. In conclusion, nnLDA performs better on learning while it requires a slightly longer training time. It is less than $10\%$ slower than DMR.

In the following section, we study the classification problem of predicting the rating of each sample. In all the cases, we use 10-fold cross validation, which holds out 10\% of the data for test purposes and trains the models on the remaining 90\%. We apply nnLDA, plain LDA and DMR to find the probability of each sample to be assigned to each topic group and treat it as the feature matrix. Lastly, we train a classification model (xgboost \cite{Chen2016XGBoostAS}) on the feature matrix with the rating labels as the ground truth.

\subsubsection{Classification}

Figures \ref{fig:proj_f1}, \ref{fig:whatif_f1} and \ref{fig:TV_f1} depict the relative F1 scores of DMR and nnLDA with respect to plain LDA on the PTS, WIP, and DCL datasets, respectively. In Figure \ref{fig:proj_f1}, the most distinguishable difference of F1 scores occurs when the number of topic groups is 15, where nnLDA has a gap of 0.032. In the meanwhile, DMR achieves its best performance at the same point with a gap of 0.030. Moreover, this chart shows that nnLDA outperforms plain LDA and DMR no matter what the number of topic groups is. In Figure \ref{fig:whatif_f1}, when using the single feature (sector attribution), the biggest gaps of F1 scores happen when the number of topic groups is 15 for DMR and 25 for nnLDA. The biggest gap between nnLDA and plain LDA is 0.016, while the largest gap between DMR and plain LDA is 0.003. Considering models using two features (sector and channel attributions) as the side data, the highest relative F1 score given by nnLDA with two features is 0.022 with 15 topic groups, compared with 0.004 produced by DMR with 10 topic groups. Although plain LDA provides a slightly higher F1 score than nnLDA when applying 5 topic groups, nnLDA outperforms plain LDA and DMR significantly given any other number of topic groups. In Figure \ref{fig:TV_f1}, the highest relative F1 score given by nnLDA is 0.022 with 25 topic groups, compared with 0.003 given by DMR for 6 topic groups. Moreover, this figure shows that nnLDA outperforms plain LDA dramatically whatever the number of topic groups is. 

Therefore, nnLDA performs better than plain LDA and DMR when predicting the rating given customer’s comments and product information in all datasets.

\subsubsection{Comment Generation}
\begin{figure*}[!t]
\centering
\begin{minipage}{0.45\textwidth}
  \centering
  \includegraphics[width=\linewidth]{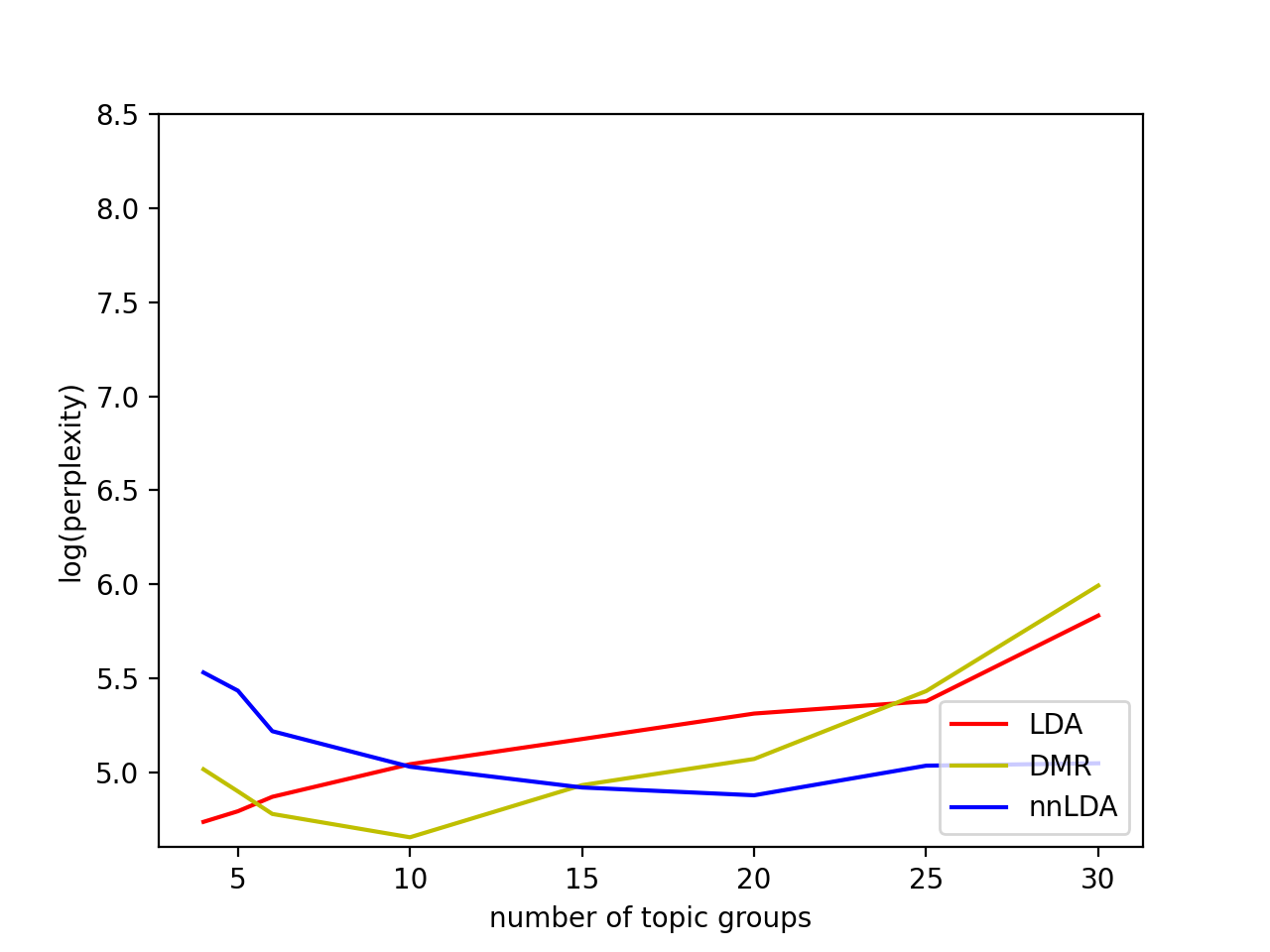}
  \caption{PTS dataset}
  \label{fig:proj_perp}
  
  \includegraphics[width=\linewidth]{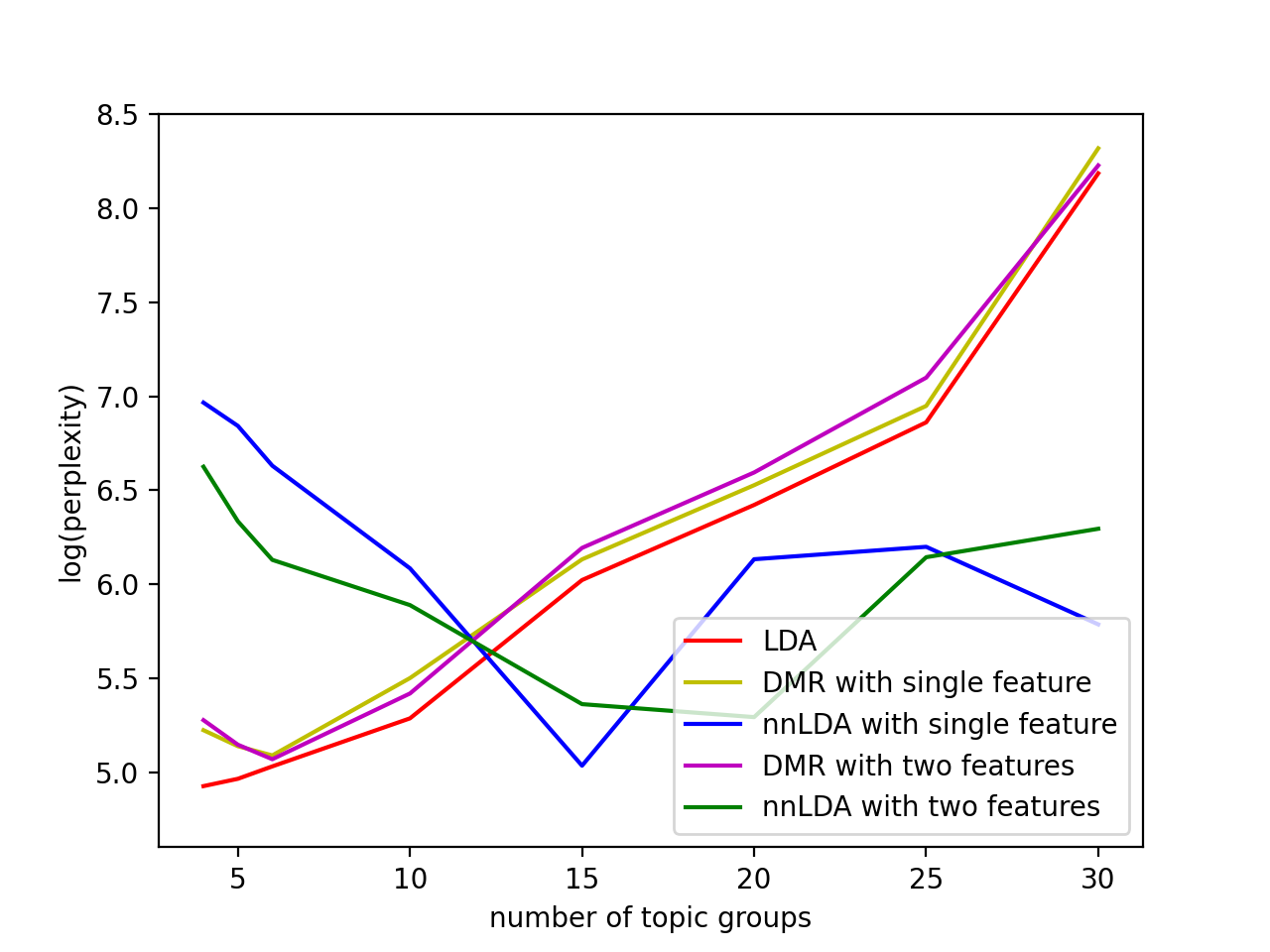}
  \caption{WIP dataset}
  \label{fig:whatif_perp}

  \includegraphics[width=\linewidth]{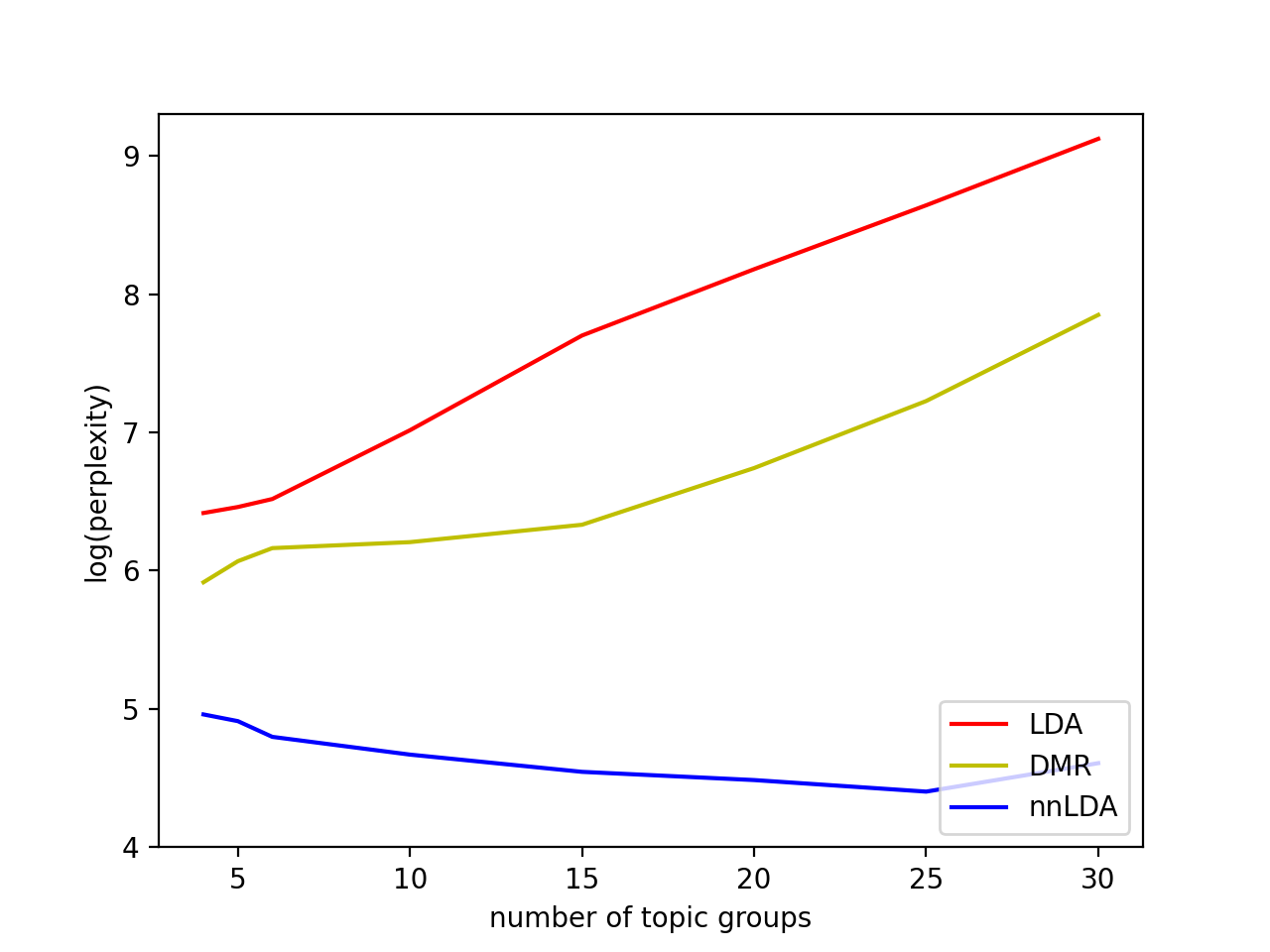}
  \caption{DCL dataset}
  \label{fig:TV_2}
\end{minipage}
\hfill
\begin{minipage}{0.45\textwidth}
  \centering
  \includegraphics[width=\linewidth]{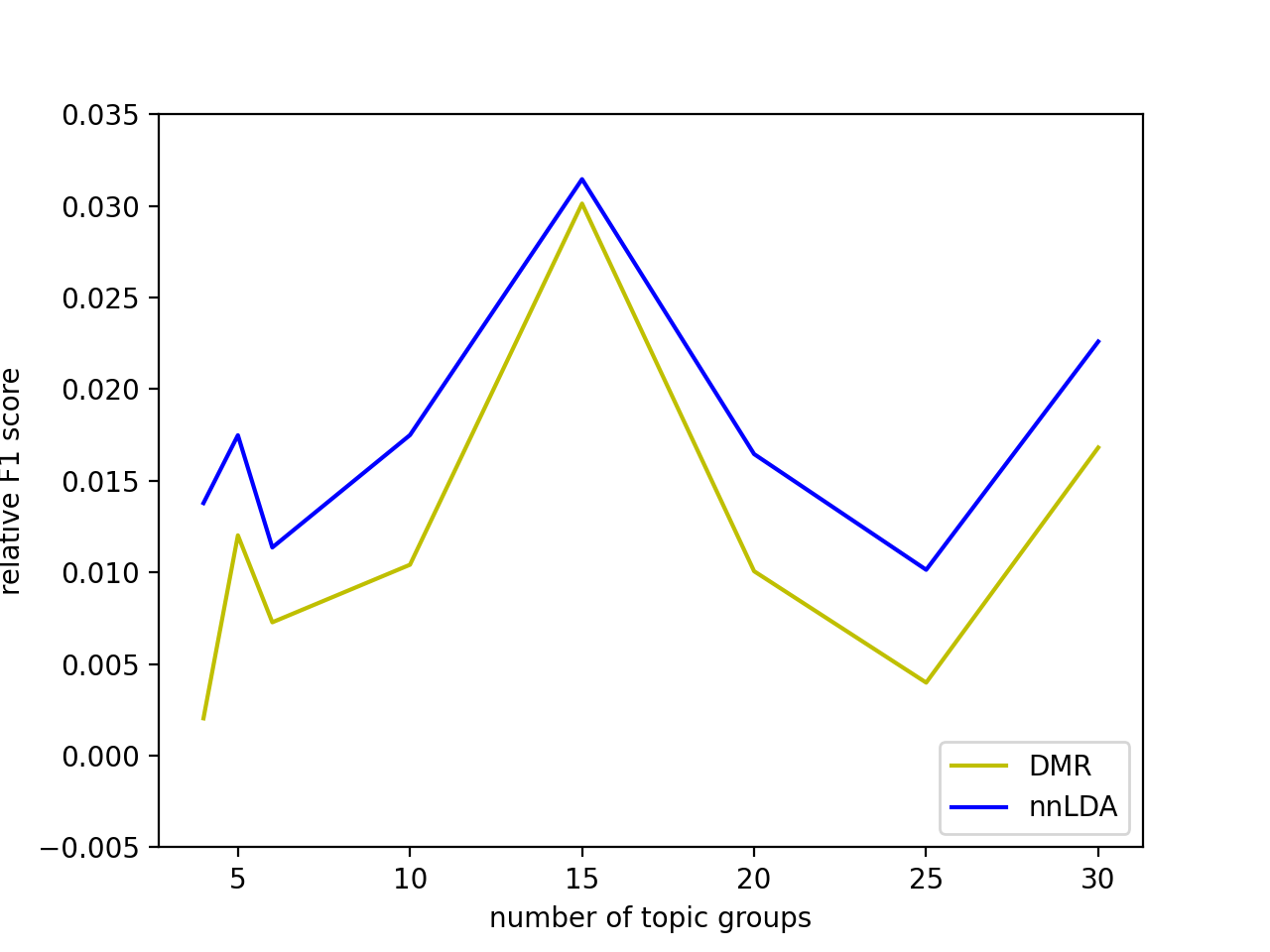}
  \caption{PTS F1 score}
  \label{fig:proj_f1}
  
  \includegraphics[width=\linewidth]{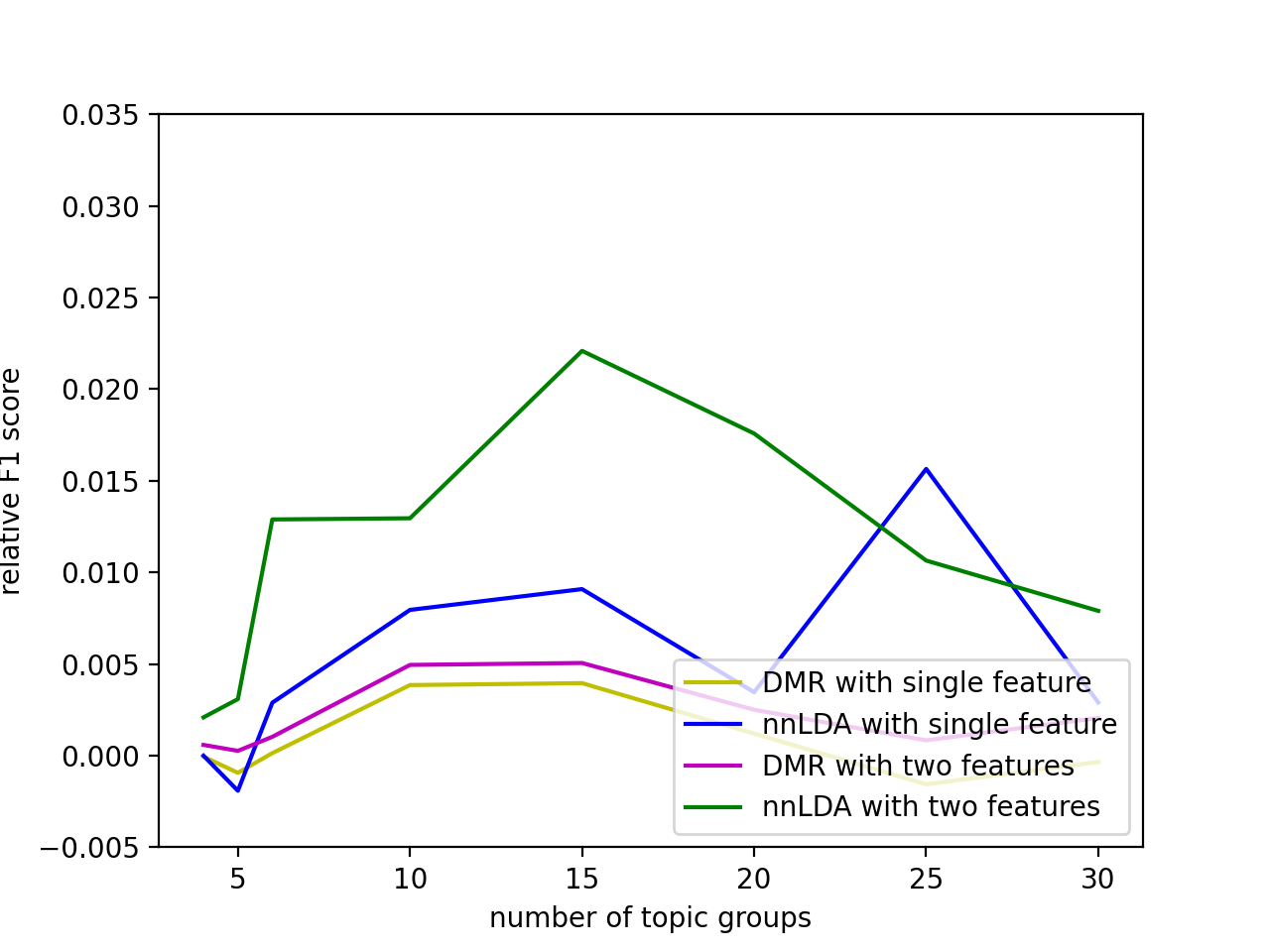}
  \caption{WIP F1 score}
  \label{fig:whatif_f1}

  \includegraphics[width=\linewidth]{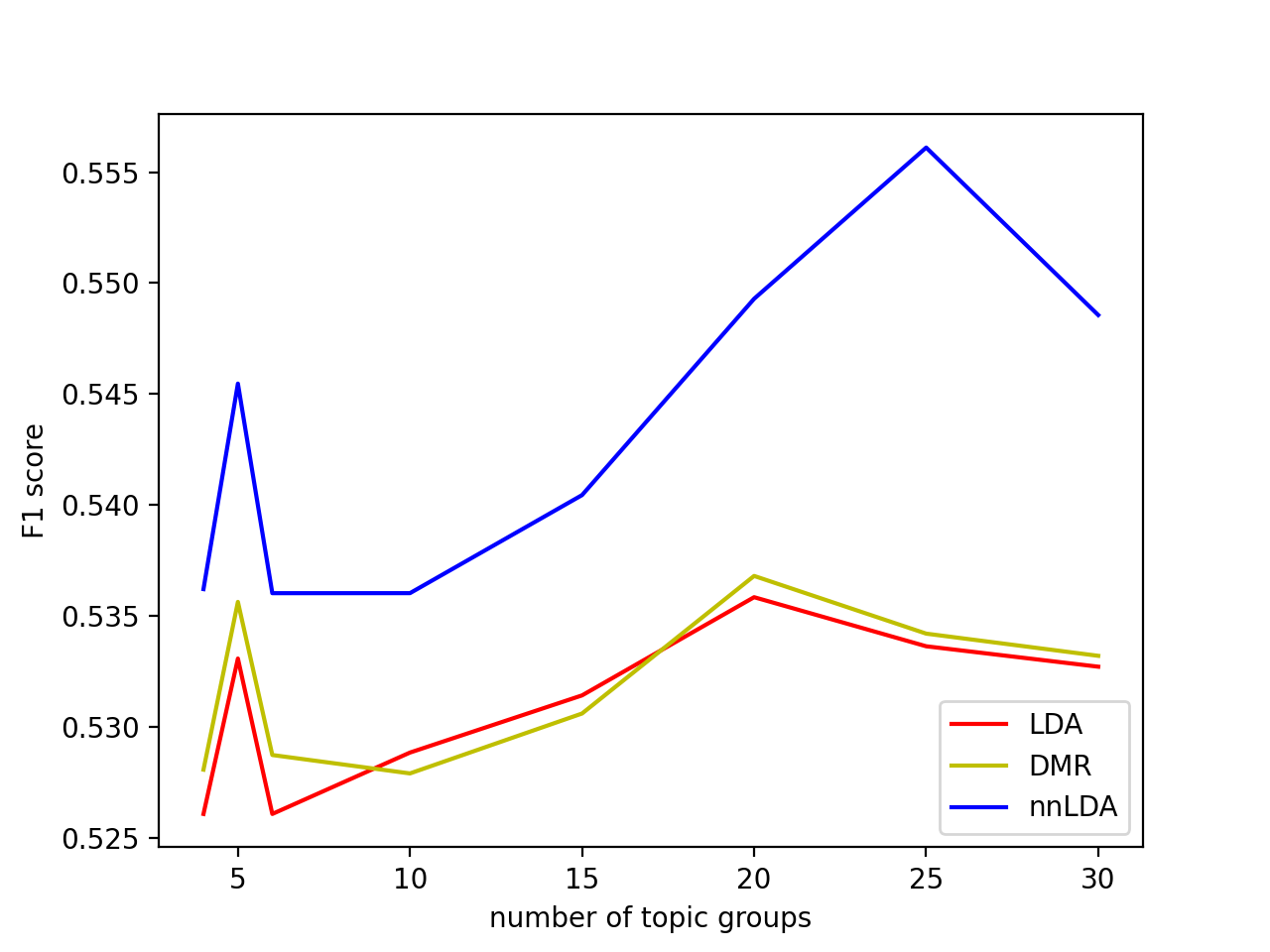}
  \caption{DCL F1 score}
  \label{fig:TV_f1}
\end{minipage}
\end{figure*}

In this section, we compare the comments generated by nnLDA with plain LDA and DMR. We set the number of topic groups to be 5 since all of plain LDA, DMR and nnLDA have relatively low perplexity scores based on Figures \ref{fig:proj_perp} and \ref{fig:whatif_perp}, and comparable F1 scores based on Figures \ref{fig:proj_f1} and \ref{fig:whatif_f1} on the PTS and WIP datasets. A comment is generated based on the topic-document probability of the sample and the topic-word distribution. More precisely, for DMR and LDA, the prior $\alpha$ is generated based on the side data (sector) first while $\alpha$ is fixed in plain LDA. Next, a comment is created by selecting the top words which have the highest score computed by adding the products of the topic-document probability and topic-word for each word. Then, we randomly pick 50 comments that contain a certain level of information, for example, we rule out comments like \say{N/A.} Meanwhile, in order to evaluate the quality of comment generation, we employed 50 PhD students. Each one of them assessed a pair of comments (one based on plain LDA or DMR, and the other one based on nnLDA) for the same side data and they provided an assessment as to which one is better.


\FloatBarrier
\FloatBarrier

The upper left three values in Table \ref{tb10:word_generate} show the comparison of the generated comments given by plain LDA and nnLDA on the PTS dataset. Based on the table, among all these 50 samples, nnLDA generates more accurate comments in 15 samples, while plain LDA does better in 11 samples, and the two are tied for the remaining 24 samples. The lower left three values in Table \ref{tb10:word_generate} show the comparison of the generated comments given by DMR and nnLDA on the PTS dataset. Based on the table, among all these 50 samples, nnLDA generates more accurate comments in 16 samples, while DMR does better in 11 samples, and the two are tied for the remaining 23 samples. On the PTS dataset, nnLDA generates in $\frac{15-11}{50}=8\%$ more reasonable comments compared to plain LDA, and in $\frac{16-11}{50}=10\%$ more comparing to DMR.

\begin{table}[t]
\centering
\resizebox{0.7\columnwidth}{!}{
\begin{tabular}{|l|r|r|}
\hline
Comparison& PTS& WIP\\
\hline
plain LDA $<$ nnLDA       & 15 & 22 \\
plain LDA $>$ nnLDA       & 11 & 9  \\
plain LDA $\approx$ nnLDA & 24 & 19 \\
\hline
DMR $<$ nnLDA             & 16 & 20 \\
DMR $>$ nnLDA             & 11 & 10 \\
DMR $\approx$ nnLDA       & 23 & 20 \\
\hline
\end{tabular}
}
\caption{Comparison of the number of generated comments on different datasets}
\label{tb10:word_generate}
\end{table}

The right column in Tables \ref{tb10:word_generate} shows the comparison of the generated comments given by plain LDA and nnLDA, and DMR and nnLDA on the WIP dataset, respectively. The observations and conclusions are similar. Furthermore, the advantage in number is more obvious on the WIP dataset, i.e. the improvement of nnLDA compared to plain LDA is as large as $\frac{22-9}{50}=26\%$ and the improvement from DMR to nnLDA is $\frac{20-10}{50}=20\%$. Therefore, taking generated comments into consideration, nnLDA generates more reasonable comments than plain LDA and DMR for both small and medium-sized datasets.

\section{Conclusion}
Our experiments confirm that integrating side data via a neural network into the LDA framework can significantly improve performance on multiple tasks. In particular, nnLDA consistently achieves higher log-likelihoods, and its adaptive prior—learned directly from side data—leads to better topic grouping, lower perplexity, and enhanced classification and comment generation. Future work will explore alternative neural network architectures to better adapt to various types of side data and will extend the evaluation to a broader range of datasets. Overall, nnLDA provides a comprehensive framework for integrating auxiliary information into topic modeling, thereby offering significant improvements over existing approaches.




\newpage
\bibliographystyle{plain}
\bibliography{references}

\newpage
\onecolumn

\section*{Appendix}
\subsection*{A \tab Probability Distribution of LDA}
\label{eq:lda}
Given the generative process of LDA, which is formally presented in \cite{blei2003latent}, we obtain the marginal distribution of a document $d=\textbf{w}$ with text only as
\begin{align*}
P_2(\textbf{w}\mid\alpha,\beta)
&=\int \hat{p}(\theta\mid\alpha)\left(\prod_{n=1}^{N}\sum_{z_{k}}\hat{p}(z_{k}\mid \theta)\hat{p}(w_{n}\mid z_{k},\beta)\right)\mathrm{d}\theta\\
&=\int \hat{p}(\theta\mid\alpha)\left(\prod_{n=1}^{N}\sum_{i=1}^{K}\prod_{j=1}^V (\theta_i\beta_{ij})^{w_n^j} \right)\mathrm{d}\theta,
\end{align*}
which in turn yields
\begin{align*}
P_2(D\mid\alpha,\beta)
&=\mathbb{E}\left[\int \hat{p}(\theta_d\mid\alpha)\left(\prod_{n=1}^{N}\sum_{z_{d_k}}\hat{p}(z_{d_k}\mid \theta_d)\hat{p}(w_{d_n}\mid z_{d_k},\beta)\right)\mathrm{d}\theta_d\right]\\
&=\mathbb{E}\left[\int \hat{p}(\theta_d\mid\alpha)\left(\prod_{n=1}^{N}\sum_{i=1}^{K}\prod_{j=1}^V (\theta_i\beta_{ij})^{w_n^j} \right)\mathrm{d}\theta_d\right],
\end{align*}
where $\hat{p}(\theta_d\mid\alpha)=p(\theta_d\mid\alpha)$. 
\subsection*{B \tab Proof of Theorem \hyperref[thm:1]{1}}\label{pf:thm1}
\begin{proof}
By finite sample expressivity of $g(\gamma;\cdot)$, there exists a model with parameters $\gamma_1$ such that
\begin{align*}
g(\gamma_1;\textbf{s})=\alpha^*,
\end{align*}
which in turn yields
\begin{align*}
\tilde{\tilde{p}}(\theta\mid\textbf{s},\gamma_1)=\hat{p}(\theta\mid g(\gamma_1;\textbf{s}))=\hat{p}(\theta\mid\alpha^*).
\end{align*}
Therefore,
\begin{align*}
P_2(D\mid\alpha^*,\beta^*)=\tilde{\tilde{P}}_1(D\mid S,\gamma_1,\beta^*)=P_1(\mu^*, \sigma^*,\gamma_1,\beta^*).
\end{align*}
Since nnLDA also optimizes over the network parameter $\gamma$, we have
\begin{align*}
P_1(D\mid \mu^*, \sigma^*,\gamma^*,\beta^*)\geq P_1(D\mid\mu^*, \sigma^*,\gamma_1,\beta^*),
\end{align*}
and thus,
\begin{align*}
P_1(D\mid\mu^*, \sigma^*,\gamma^*,\beta^*)\geq P_2(D\mid\alpha^*,\beta^*).
\end{align*}
\end{proof}
\subsection*{C \tab Proof of Theorem \hyperref[thm:2]{2}}\label{pf:thm2}
\begin{proof}
Note that
\begin{align}
\label{thm:2-1}
&\frac{P_1(D\mid\mu^*,\sigma^*,\gamma^*,\beta^* )-P_2(D\mid\alpha^*,\beta^*)}{P_2(D\mid\alpha^*,\beta^*)}\nonumber\\
=&
\frac{\mathbb{E}\left[\int \tilde{p}(\theta_d\mid \mu^*,\sigma^*,\gamma^*)\left(\prod_{n=1}^{N}\sum_{z_{d_k}}\tilde{p}(z_{d_k}\mid \theta_d)\tilde{p}(w_{d_n}\mid z_{d_k},\beta^*)\right)\mathrm{d}\theta_d\right]}
{\mathbb{E}\left[\int \hat{p}(\theta_d\mid\alpha^*)\left(\prod_{n=1}^{N}\sum_{z_{d_k}}\hat{p}(z_{d_k}\mid \theta_d)\hat{p}(w_{d_n}\mid z_{d_k},\beta^*)\right)\mathrm{d}\theta_d\right]}\nonumber\\
&\quad\quad\quad-\frac{\mathbb{E}\left[\int \hat{p}(\theta_d\mid\alpha^*)\left(\prod_{n=1}^{N}\sum_{z_{d_k}}\hat{p}(z_{d_k}\mid \theta_d)\hat{p}(w_{d_n}\mid z_{d_k},\beta^*)\right)\mathrm{d}\theta_d\right]}
{\mathbb{E}\left[\int \hat{p}(\theta_d\mid\alpha^*)\left(\prod_{n=1}^{N}\sum_{z_{d_k}}\hat{p}(z_{d_k}\mid \theta_d)\hat{p}(w_{d_n}\mid z_{d_k},\beta^*)\right)\mathrm{d}\theta_d\right]}.
\end{align}
Since
\begin{align*}
&\tilde{p}(\theta_d\mid \mu^*,\sigma^*,\gamma^*)\left(\prod_{n=1}^{N}\sum_{z_{d_k}}\tilde{p}(z_{d_k}\mid \theta_d)\tilde{p}(w_{d_n}\mid z_{d_k},\beta^*)\right)\\
=&\tilde{p}(\theta_d\mid\mu^*,\sigma^*,\gamma^*)\left(\prod_{n=1}^{N}\tilde{p}(w_{d_n}\mid \theta_d,\beta^*)\right)=\prod_{n=1}^{N}\tilde{p}(w_{d_n}\mid \gamma^*,\beta^*, \mu^*,\sigma^*)
\end{align*}
and
\begin{align*}
&\hat{p}(\theta_d\mid\alpha^*)\left(\prod_{n=1}^{N}\sum_{z_{d_k}}\hat{p}(z_{d_k}\mid \theta_d)\hat{p}(w_{d_n}\mid z_{d_k},\beta^*)\right)\\
=&\hat{p}(\theta_d\mid\alpha^*)\left(\prod_{n=1}^{N}\hat{p}(w_{d_n}\mid \theta_d,\beta^*)\right)=\prod_{n=1}^{N}\hat{p}(w_{d_n}\mid \alpha^*,\beta^*),
\end{align*}
equation (\ref{thm:2-1}) could be further simplified as
\begin{align*}
&\frac{P_1(D\mid\mu^*,\sigma^*,\gamma^*,\beta^*)-P_2(D\mid\alpha^*,\beta^*)}{P_2(D\mid\alpha^*,\beta^*)}\\
=& \frac{\mathbb{E}\left[\int\prod_{n=1}^{N}\tilde{p}(w_{d_n}\mid \mu^*,\sigma^*,\gamma^*,\beta^*)\mathrm{d}\theta_d\right]-
\mathbb{E}\left[\int\prod_{n=1}^{N}\hat{p}(w_{d_n}\mid \alpha^*,\beta^*)\mathrm{d}\theta_d\right]}
{\mathbb{E}\left[\int\prod_{n=1}^{N}\hat{p}(w_{d_n}\mid \alpha^*,\beta^*)\mathrm{d}\theta_d\right]}\\
\geq& 
\frac{\mathbb{E}\left[\int C\prod_{n=1}^{N}\hat{p}(w_{d_n}\mid \alpha^*,\beta^*)\mathrm{d}\theta_d\right]-
\mathbb{E}\left[\int\prod_{n=1}^{N}\hat{p}(w_{d_n}\mid \alpha^*,\beta^*)\mathrm{d}\theta_d\right]}
{\mathbb{E}\left[\int\prod_{n=1}^{N}\hat{p}(w_{d_n}\mid \alpha^*,\beta^*)\mathrm{d}\theta_d\right]}\\
=&
\frac{C\cdot \mathbb{E}\left[\int \prod_{n=1}^{N}\hat{p}(w_{d_n}\mid \alpha^*,\beta^*)\mathrm{d}\theta_d\right]-
\mathbb{E}\left[\int\prod_{n=1}^{N}\hat{p}(w_{d_n}\mid \alpha^*,\beta^*)\mathrm{d}\theta_d\right]}
{\mathbb{E}\left[\int\prod_{n=1}^{N}\hat{p}(w_{d_n}\mid \alpha^*,\beta^*)\mathrm{d}\theta_d\right]}=C-1.
\end{align*}

\end{proof}
\end{document}